\documentclass[letterpaper, 10 pt, conference]{ieeeconf}  % Comment this line out if you need a4paper

\IEEEoverridecommandlockouts                              % This command is only needed if 
\usepackage{amsmath}
\usepackage{graphicx}

\usepackage{subcaption}
\usepackage[percent]{overpic} % 用于在图片上覆盖文字
\usepackage{xcolor}          % 处理颜色命令
\usepackage{array}
\usepackage{url}
\usepackage{hyperref}
\title{\LARGE \bf
PRIOR: Perceptive Learning for Humanoid Locomotion with Reference Gait Priors\vspace{-0.1cm}
}

\author{
Chenxi Han$^{1,2,*}$,
Shilu He$^{2,*}$,
Yi Cheng$^{2}$,
Linqi Ye$^{3}$,
Houde Liu$^{1,\dagger}$\\
$^{1}$ Tsinghua University \quad
$^{2}$ ZERITH Robotics  \quad
$^{3}$ Shanghai University \\
$^{*}$ Equal contribution \quad
$^{\dagger}$ Corresponding author \vspace{-0.4cm}
}

\begin{document}
\maketitle
\thispagestyle{empty}
\pagestyle{empty}

%%%%%%%%%%%%%%%%%%%%%%%%%%%%%%%%%%%%%%%%%%%%%%%%%%%%%%%%%%%%%%%%%%%%%%%%%%%%%%%%
\begin{abstract}
%V1
% Humanoid robots operating in complex real-world environments must achieve strong terrain adaptability while maintaining natural, human-like locomotion. However, existing approaches often struggle to reconcile perception-driven robustness with motion-prior-driven human-likeness. This paper presents \textbf{PRIOR}, a locomotion framework that unifies perceptual learning with reference gait priors. On the perception side, PRIOR incorporates an efficient GRU-based state estimator that explicitly predicts linear velocity via self-supervised learning while implicitly inferring future proprioceptive states, and captures essential terrain geometry through an elevation reconstruction task. On the motion generation side, parameterized reference trajectories are constructed from high-fidelity motion capture data and integrated into policy optimization via gait-phase and joint-level priors, enabling a deep coupling between perception and motion structure. Experimental results demonstrate that PRIOR achieves highly human-like gait patterns and strong dynamic robustness in both simulation and on the custom-built humanoid prototype ZERITH Z1. Successful deployment on minimally calibrated hardware highlights the robustness and generalization capability of the proposed framework. The framework will be released to support future research.
%V2
Training perceptive humanoid locomotion policies that traverse complex terrains with natural gaits remains an open challenge, typically demanding multi-stage training pipelines, adversarial objectives, or extensive real-world calibration. We present PRIOR, an efficient and reproducible framework built on Isaac Lab that achieves robust terrain traversal with human-like gaits through a simple yet effective design: (i) a parametric gait generator that supplies stable reference trajectories derived from motion capture without adversarial training, (ii) a GRU-based state estimator that infers terrain geometry directly from egocentric depth images via self-supervised heightmap reconstruction, and (iii) terrain-adaptive footstep rewards that guide foot placement toward traversable regions. Through systematic analysis of depth image resolution trade-offs, we identify configurations that maximize terrain fidelity under real-time constraints, substantially reducing perceptual overhead without degrading traversal performance. Comprehensive experiments across terrains of varying difficulty—including stairs, boxes, and gaps—demonstrate that each component yields complementary and essential performance gains, with the full framework achieving a 100\% traversal success rate. We will open-source the complete PRIOR framework, including the training pipeline, parametric gait generator, and evaluation benchmarks, to serve as a reproducible foundation for humanoid locomotion research on Isaac Lab.
\end{abstract}

%%%%%%%%%%%%%%%%%%%%%%%%%%%%%%%%%%%%%%%%%%%%%%%%%%%%%%%%%%%%%%%%%%%%%%%%%%%%%%%%
\section{INTRODUCTION}
Deploying humanoid robots in human-centric environments requires locomotion controllers that can negotiate diverse terrain geometries—stairs, boxes, gaps—while preserving the natural bipedal gaits essential for safe and predictable coexistence with people and infrastructure.
Reinforcement learning (RL) has emerged as the dominant paradigm for training such controllers, with recent work demonstrating impressive results in either terrain-robust locomotion through perceptive policies~\cite{zhuang2024humanoid, 10678805} or natural gait synthesis through motion priors~\cite{peng2020learning,Peng_2022}.
Achieving both capabilities simultaneously, however, typically incurs substantial system complexity: multi-stage teacher--student distillation, adversarial discriminators for style enforcement, or extensive sim-to-real calibration.
These requirements hinder reproducibility and raise the barrier to entry for locomotion research.
In this work, we ask: \emph{can a single-stage RL pipeline, without adversarial training or distillation, produce perceptive humanoid locomotion that is both terrain-robust and natural?}

%%%%%%%%%%%%%%%%%%%%%%%%%%%%%%%%%%%%%%%%%%%%%%%%%%%%%%%%%%%%%%%%%%%
\begin{figure}[t]
   \centering
   \includegraphics[width=1.0\columnwidth]{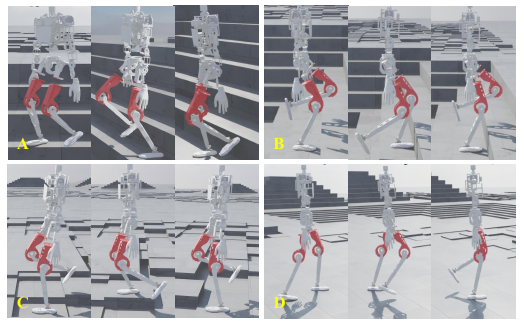}
   \caption{The proposed PRIOR framework was simulated and demonstrated on the ZERITH Z1 model. (A)--(D) illustrate the robot traversing four representative terrain types.}
   \label{fig:walk_fig}
\end{figure}
%%%%%%%%%%%%%%%%%%%%%%%%%%%%%%%%%%%%%%%%%%%%%%%%%%%%%%%%%%%%%%%%%%%%%%%%%%%

We answer this question with \textbf{PRIOR}, a framework whose design is guided by three observations drawn from recent literature.

\textbf{(i)}~LiDAR-based elevation mapping, while geometrically precise, relies on odometric integration that accumulates drift during extended locomotion; egocentric depth images, by contrast, provide a self-contained terrain signal that is inherently free of such drift.

\textbf{(ii)}~Adversarial motion priors, the prevailing mechanism for imposing gait style, suffer from well-documented training pathologies—mode collapse, reward ambiguity, and hyperparameter sensitivity — that are exacerbated on challenging terrains where the policy must deviate significantly from reference motions; a parametric gait generator can offer comparable stylistic guidance through deterministic supervision, sidestepping these instabilities entirely.

\textbf{(iii)}~The ongoing transition from Isaac Gym to Isaac Lab as the community-standard simulation platform calls for training pipelines natively built on the newer stack; we develop PRIOR entirely within Isaac Lab, incorporating systematic optimizations that yield a $3\times$ speedup over the vanilla Isaac Lab baseline.

Concretely, PRIOR trains a single locomotion policy end-to-end through three mutually reinforcing mechanisms.
A \emph{parametric gait generator} produces phase-conditioned joint trajectories by dynamically blending motion capture primitives, supplying the policy with velocity-adaptive motion targets that replace adversarial style losses.
A \emph{GRU-based state estimator} fuses proprioceptive history with egocentric depth observations and is trained via self-supervised auxiliary objectives—heightmap reconstruction and linear velocity prediction—to distill local terrain geometry into a compact latent representation, requiring neither external localization nor manual annotation.
\emph{Terrain-adaptive footstep rewards} bias swing-leg placement toward geometrically favorable contact regions, enabling reliable footholds on discontinuous surfaces.
Complementing these components, we conduct a depth resolution study that characterizes the Pareto frontier between reconstruction fidelity and computational cost, revealing that perceptual overhead can be substantially reduced with negligible impact on traversal performance.

We validate PRIOR on simulated terrains of progressive difficulty—flat ground, boxes, staircases, and gaps.
Controlled ablations confirm that each component is individually necessary and that their combination produces synergistic gains beyond any subset, with the complete system reaching a 100\% traversal success rate.
To lower the barrier for future work, we will publicly release the full code.
Our contributions are summarized as follows:

\begin{itemize}
  \item We present PRIOR, a single-stage RL framework that unifies depth-based terrain perception, parametric gait generation, and terrain-adaptive footstep rewards to achieve robust and natural humanoid locomotion—eliminating the need for adversarial objectives, teacher--student distillation, or multi-stage training.

 \item We provide a depth resolution analysis that maps the Pareto frontier between terrain reconstruction quality and computational throughput, together with Isaac Lab-specific training optimizations that yield a $3\times$ training speedup over the vanilla implementation.

  \item We conduct comprehensive ablations that isolate the necessity and synergy of each component, and commit to open-sourcing the complete framework as a reproducible baseline on Isaac Lab.
\end{itemize}

\section{RELATED WORK}

\subsection{Perception-Driven Robot Locomotion}
Early research on legged robot locomotion focused primarily on "blind walking" strategies based on proprioception. By constructing closed-loop feedback control using internal sensors such as joint encoders and IMUs, these methods achieved a degree of robust walking on unknown terrains. Studies such as \cite{10161144, KumarA-RSS-21, Lee_2020, Ji_2022, Hwangbo_2019} have demonstrated that strong terrain adaptability can be attained relying solely on internal sensing. Based on this, several works have introduced motion prior constraints to generate more agile and holistic locomotion patterns \cite{10167753, wu2023learning}. However, due to the lack of direct environmental observation, these methods struggle with precise, proactive gait planning when encountering significant non-local terrain variations.

As task complexity has increased, recent research has gradually shifted toward end-to-end locomotion learning frameworks. One category of methods introduces two-stage teacher-student distillation frameworks with privileged information  \cite{zhuang2024humanoid, agarwal2022legged, 10610200, wang2025beamdojo}. Some studies have incorporated temporal structures and attention mechanisms to enhance the robustness and consistency of terrain representations \cite{10678805, doi:10.1126/scirobotics.adv3604, sun2025dpl}, while some other work has explored the integration of gait optimization \cite{wang2025more}. Despite the significant progress made by these methods in complex environments, their motion generation often relies heavily on meticulous reward engineering or multi-stage training pipelines. This results in high system complexity and leaves room for improvement regarding the naturalness of the generated motion.

\subsection{Motion-Prior-Based Robot Locomotion}
Research on motion priors primarily focuses on enhancing the human-likeness of robot movements and can be broadly categorized into explicit motion generation and stylistic constraints. One prominent category follows the imitation learning paradigm. Some approaches achieve complex skill synthesis and multi-action switching through explicit trajectory constraints \cite{10.1145/3197517.3201311, 10.1145/3592447}. Others utilize latent variables or probabilistic models to model high-dimensional motion distributions in order to generate diverse human-like movements  \cite{Peng_2022, 11247682}. Recently, diffusion models have emerged as a significant tool for constructing motion priors, demonstrating remarkable advantages \cite{10.1145/3680528.3687626, liao2025beyondmimic}. Despite significantly improving gait naturalness, insufficient modeling of environmental perception and online feedback limits its adaptability and robustness in complex terrain.

Another category of methods utilizes reference motions as reward signals or discriminative criteria to guide the policy toward human-like styles in a weakly supervised manner. This approach enhances motion flexibility by relaxing strict trajectory matching constraints. Representative works include adversarial imitation for physical character control \cite{Peng_2021} and its applications in legged robot locomotion control \cite{10167753, 10.1007/978-3-031-72062-8_11, shi2025adversarial}. Furthermore, some studies have utilized latent space modeling or Mixture-of-Experts (MoE) structures to enhance gait diversity while improving the continuity of motion generation \cite{wu2023learning, wang2025more}. However, due to insufficient integration of environmental perception, learned human-like styles struggle to maintain consistency in real-world or unstructured environments.

\section{METHOD}
In this section, we provide a detailed description of the implementation of the PRIOR framework. PRIOR utilizes an estimator that fuses temporal depth information with proprioception to estimate both terrain features and robot proprioceptive states. Furthermore, reference gaits derived from processed natural motion data are integrated as inputs, providing soft constraints on robot locomotion via gait-aware rewards. Our discussion is organized into four primary components: the overall perceptive locomotion framework, the generation and learning of human-like motions, the high-throughput training infrastructure, and the specific implementation details of the training process.

\begin{figure*}[t]
   \centering
   \includegraphics[width=1.0\textwidth]{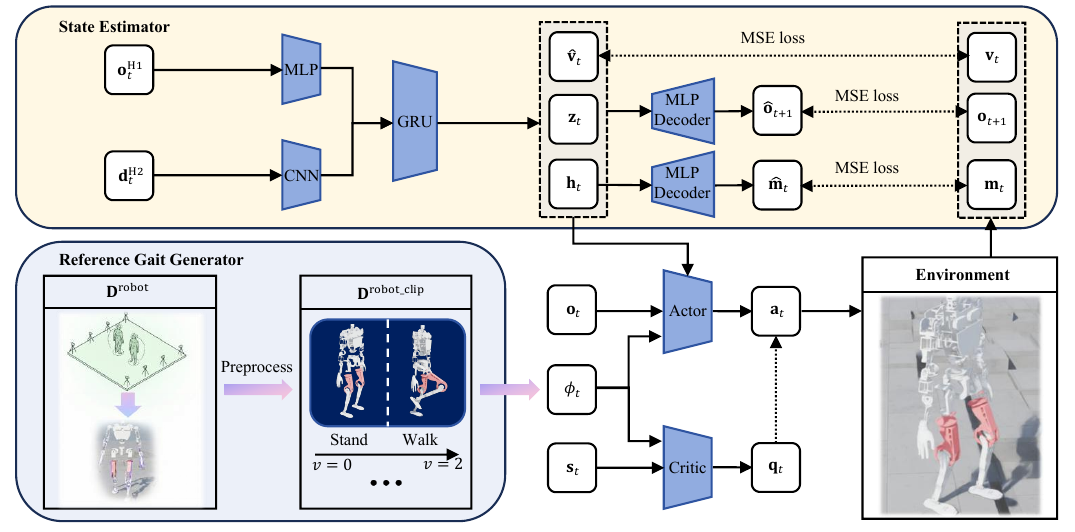}
   \caption{Overview of the proposed PRIOR framework. The framework comprises three components: (a) Asymmetric actor-critic architecture for reinforcement learning. (b) \textbf{State Estimator} (yellow): Fuses multimodal latent features for policy driving and performs self-supervised regression for velocity estimation, terrain reconstruction, and state prediction. (c) \textbf{Reference Gait Generator} (blue): Synthesizes physics-consistent reference trajectories via phase normalization and velocity-driven weighted interpolation, while constraining locomotion through gait-aware rewards.}
   \label{fig:network_fig}
\end{figure*}

\subsection{Perception-Driven Locomotion Framework}
\subsubsection{Asymmetric Actor–Critic Architecture}
As illustrated in Fig.~\ref{fig:network_fig}, the proposed PRIOR framework adopts an asymmetric actor–critic architecture and is trained in a single-stage, end-to-end manner that enables the perception module and the control policy to co-evolve \cite{10678805}. This design avoids error amplification issues commonly encountered in two-stage teacher–student distillation-based training pipelines. The policy is optimized using Proximal Policy Optimization (PPO) \cite{schulman2017proximal}.

\textbf{Actor network.} The actor network takes (i) a 45-dimensional current proprioceptive observation $\mathbf{o}_t$, (ii) a 163-dimensional state estimator output $\mathbf{e}_t$ which is detailed in Section~\ref{sec:state_estimator}, and (iii) a 1-dimensional gait phase signal $\phi_{t}$ as input. The current proprioceptive observation is defined as follows:
\begin{equation}
\mathbf{o}_t =
\left[
\boldsymbol{\omega}_t,\;
\mathbf{g}_t,\;
\mathbf{c}_t,\;
\boldsymbol{\theta}_t,\;
\dot{\boldsymbol{\theta}}_t,\;
\mathbf{a}_{t-1}
\right]^{\top}.
\end{equation}
where $\boldsymbol{\omega}_t$ is the body angular velocity, $\mathbf{g}_t$ is the gravity direction vector expressed in the body frame, $\mathbf{c}_t$ is the velocity command, $\boldsymbol{\theta}_t$ and $\dot{\boldsymbol{\theta}}_t$ are the joint positions and velocities, respectively, and $\mathbf{a}_{t-1}$ is the action applied at
the previous time step.

\textbf{Critic Network.} The critic network receives the noise-free base linear velocity $\mathbf{v}_t$ from the simulation environment, the proprioceptive observation $\mathbf{o}_t$, the height map scan information $\mathbf{m}_t$ provided by the RayCaster, and the reference gait phase $\phi_t$. The input is defined as:
\begin{equation}
\mathbf{s}_t =
\left[
\mathbf{v}_t,\;
\mathbf{o}_t,\;
\mathbf{m}_t\;
\right]^{\top}.
\end{equation}

\textbf{Reward and Action Space.} We categorize the reward functions into four groups: task tracking, stability, smoothness, and safety. Specific rewards for foot-terrain interaction are crucial for humanoid balance and obstacle negotiation. These foot-related components, summarized in Table \ref{tab:common_rewards}, ensure gait rhythm, ground clearance, and precise landing on complex environmental features.

The actor network outputs a 12-dimensional action vector $\mathbf{a}_t$ corresponding to the leg joints of the humanoid robot. This action serves as modulation to the default standing joint configuration $\boldsymbol{\theta}^{\mathrm{default}}$, yielding the target joint positions $\boldsymbol{\theta}_{t}^{\mathrm{target}}$,which is defined as:
\begin{equation}
\boldsymbol{\theta}_{t}^{\mathrm{target}} = \boldsymbol{\theta}^{\mathrm{default}} + \mathbf{a}_t.
\end{equation}
The desired joint positions are then fed into a low-level PD controller to calculate the target joint torques $\boldsymbol{\tau}_t$, which is defined as:
\begin{equation}
\boldsymbol{\tau}_t = K_p(\boldsymbol{\theta}_{t}^{\mathrm{target}} - \boldsymbol{\theta}_t) - K_d\dot{\boldsymbol{\theta}}_t.
\end{equation}
where the stiffness $K_p$ and damping $K_d$ are set to 60.0 and 2.0, respectively, to match the hardware specifications of ZERITH Z1.

\begin{table}[t]
\caption{LANDING STATE REWARD COMPONENTS}
\label{tab:common_rewards}
\centering
\renewcommand{\arraystretch}{1.3}
\begin{tabular*}{\columnwidth}{@{\extracolsep{\fill}}lcc}
\hline
\textbf{Reward} & \textbf{Description} & \textbf{Weight ($w_i$)} \\
\hline
$r_{\text{air}}$     & Promotes gait rhythm.               & 1.25 \\
$r_{\text{slide}}$   & Minimizes ground slipping.          & -0.10 \\
$r_{\text{dbl-air}}$ & Penalizes walking on one leg.       & -1.00 \\
$r_{\text{swing}}$   & Ensures leg lift height.            & -20.0 \\
$r_{\text{stumble}}$ & Prevents foot-obstacle tripping.    & -30.0 \\
$r_{\text{edge}}^{L/R}$ & Encourages safe foot placement.  & -2.00 \\
\hline
\end{tabular*}
\end{table}

\subsubsection{State Estimator}
\label{sec:state_estimator}
The state estimator fuses proprioceptive sensing and depth visual perception. Compared with estimation methods that rely solely on visual information, this multimodal fusion architecture improves the real-time performance of state feedback and reduces estimation bias caused by noise in a single visual modality. The two input modalities consist of proprioceptive observations $\mathbf{o}_t^{\mathrm{H1}}$ with a stacking horizon of H1 = 10, and temporally stacked depth images $\mathbf{d}_t^{\mathrm{H2}}$ with a stacking horizon of H2 = 2, where each depth frame has a cropped resolution of [36,64]. The proprioceptive observation $\mathbf{o}_t^{\mathrm{H1}}$ is processed by a multilayer perceptron (MLP) encoder to extract a 128-dimensional proprioceptive state feature, while the temporally stacked depth images $\mathbf{d}_t^{\mathrm{H2}}$ are processed by a convolutional neural network (CNN) encoder to extract a 128-dimensional depth feature. The encoded features of the two modalities are then concatenated and fed into a single-layer gated recurrent unit (GRU) to generate a memory representation of the proprioceptive state and the terrain state. The output of the memory module is a 163-dimensional vector $\mathbf{e}_t$, which serves as an input to the actor network, composed of a 3-dimensional base velocity $\hat{\mathbf{v}}_t$, a 32-dimensional latent vector $\mathbf{z}_t$ and a 128-dimensional height map latent vector $\mathbf{h}_t$.
\begin{equation}
\mathbf{e}_t =
\left[
\hat{\mathbf{v}}_t,\;
\mathbf{z}_t,\;
\mathbf{h}_t
\right]^{\top}.
\end{equation}
The height map latent vector $\mathbf{h}_t$ is decoded by an MLP to obtain the estimated terrain state $\hat{\mathbf{m}}_t$, while $\mathbf{e}_t$ is decoded by another MLP to predict the next-step proprioceptive state $\hat{\mathbf{o}}_{t+1}$. Using the privileged information available in the critic observations, self-supervised learning is applied to $\hat{\mathbf{v}}_t$, as well as the decoded results $\hat{\mathbf{m}}_t$ and $\hat{\mathbf{o}}_{t+1}$, by minimizing the mean squared error (MSE) loss. The overall loss function of the state estimator is defined as follows:
\begin{equation}
\mathcal{L} = 
\mathrm{MSE}\!\left( \hat{\mathbf{v}}_t, \mathbf{v}_t \right)
+ \mathrm{MSE}\!\left( \hat{\mathbf{o}}_{t+1}, \mathbf{o}_{t+1} \right)
+ \mathrm{MSE}\!\left( \hat{\mathbf{m}}_t, \mathbf{m}_t \right)
\end{equation}

\subsection{Reference Gait Priors}
\label{sec:ref_gait_priors}
Inspired by the Parameterized Motion Generator (PMG) framework \cite{han2026pmgparameterizedmotiongenerator}, we present a data-efficient motion prior extraction method designed to achieve humanoid control on complex terrains using a minimal set of high-fidelity motion templates. Distinguishing itself from conventional imitation learning approaches that necessitate hours of large-scale motion capture data, our method extracts structural features from core gait cycles to construct a continuous and physics-consistent reference gait space. This not only substantially lowers the overhead of data acquisition and preprocessing but also provides robust kinematic guidance for policy convergence in non-stationary terrain environments.
\subsubsection{Motion Data Preprocessing}
We utilize a high-precision motion capture system to collect human movement data, ranging from static postures to various forward velocities, denoted as $\mathbf{D}^{\mathrm{human}}$. This data is then retargeted to our humanoid robot, ZERITH Z1, using optimization-based algorithms \cite{10801984}, \cite{araujo2025gmr}, resulting in the robot dataset $\mathbf{D}^{\mathrm{robot}}$:
\begin{equation}
\mathbf{D}^{\mathrm{robot}} = 
\{ \boldsymbol{\theta}, \dot{\boldsymbol{\theta}}, \mathbf{v}, \boldsymbol{\omega}, \mathbf{c} \}
\end{equation}
$\boldsymbol{\theta}$ represents the joint angles, $\dot{\boldsymbol{\theta}}$ represents the joint angular velocities, $\mathbf{v}$ represents the base linear velocity, $\boldsymbol{\omega}$ represents the base angular velocity, $\mathbf{c} = \{\mu,\sigma\}$ represents the foot contact information, including the contact center and range.

Since the raw retargeted dataset $\mathbf{D}^{\mathrm{robot}}$ contains initial transitions, terminal decelerations, and measurement noise inherent in the motion capture process, we designed a preprocessing pipeline to extract high-fidelity, periodic, and smooth reference motion segments. To construct reference trajectories with strict periodicity, we utilize the foot contact information $\mathbf{c}$ to segment the motion data at various velocities. We employ a clip\_range mechanism to extract a single, stable gait cycle $T$ from the original long sequences. In practice, we typically select the second or third cycle from the sequence to avoid the non-stationary dynamics associated with acceleration or deceleration phases. Furthermore, to eliminate high-frequency noise introduced by the motion capture system and ensure the smoothness of joint commands, we apply a 1D Gaussian filter to the raw joint sequences. The resulting processed robot dataset is represented as:
\begin{equation}
\mathbf{D}^{\mathrm{robot\_clip}} = 
\{ \boldsymbol{\theta}, \dot{\boldsymbol{\theta}}, \mathbf{v}, \boldsymbol{\omega}, \mathbf{c}, T\}
\end{equation}
where T denotes the duration of a single gait cycle.

\subsubsection{Reference Gait Generation}
Given the commanded base velocity 
$\mathbf{v} = [v_x, v_y, \omega]$ 
and the gait phase $\phi \in [0,1)$, 
the reference joint trajectory is synthesized through a unified weighted interpolation framework.

For each velocity channel $x \in \{v_x, v_y, \omega\}$, 
we determine an interpolation coefficient based on the magnitude of the commanded velocity. 
Assume that the dataset contains a set of motion templates 
$\{\theta_i(\phi), T_i\}$, 
where $\theta_i(\phi)$ denotes the phase-dependent joint trajectory 
and $T_i$ is the corresponding gait period.

Given a commanded velocity $u_x$, 
we select its two neighboring nominal velocities 
$u_l$ and $u_u$, and define the interpolation factor as
\begin{equation}
\alpha = \mathrm{clip} \left( \frac{|u_x| - u_l}{u_u - u_l + \varepsilon}, 0,1
\right)
\end{equation}
where $\varepsilon$ is a small constant for numerical stability.

The commanded gait period is obtained via linear interpolation:
\begin{equation}
T_u = (1-\alpha) T_l + \alpha T_u .
\end{equation}
The gait phase is then updated according to the normalized phase progression rule:
\begin{equation}
\phi_{t+1} = \left( \phi_t + \frac{\Delta t}{T_u} \right)
\bmod 1 .
\end{equation}
After obtaining the updated phase $\phi$, 
the reference joint trajectory is synthesized by blending 
the neighboring motion templates:
\begin{equation}
\theta_d(\phi) = (1-\alpha)\, \theta_l(\phi) + \alpha\, \theta_u(\phi).
\end{equation}
To handle near-zero velocity commands, 
we introduce a standing threshold $v_{\text{th}}$:
\begin{equation}
\theta_d(\phi) = 
\begin{cases}
\theta_{\text{stand}}, & \|v\| \le v_{\text{th}}, \\
(1-\alpha) \theta_l(\phi) + \alpha \theta_u(\phi), & \text{otherwise}.
\end{cases}
\end{equation}

In addition, a velocity-dependent stance ratio $\rho(\mathbf{v})$ 
is defined to construct phase-based contact indicators 
$r^{L}(\phi)$ and $r^{R}(\phi)$, 
which are later used for contact supervision and reward formulation.

This velocity-conditioned interpolation strategy ensures smooth transitions across different commanded speeds and maintains temporal consistency via unified phase evolution.

\subsubsection{Gait-aware Reward Design}
To encourage the policy to maintain consistency with the reference gait over complex terrains, 
we construct a set of exponential tracking reward terms based on the target joint trajectories 
and commanded base velocities generated by the reference gait module. 
Each reward term follows a unified exponential form:
\begin{equation}
r_i = \exp(-\lambda_i e_i),
\end{equation}
where $e_i$ denotes the tracking error of the corresponding physical quantity, 
and $\lambda_i$ is a scaling coefficient.
All gait-related reward terms are combined as a weighted summation:
\begin{equation}
r_{\text{gait}} = \sum_i w_i r_i,
\end{equation}
where $w_i$ represents the weight of each component.

The detailed definitions of each gait reward term are summarized in Table \ref{tab:gait_rewards}.

These reward components constrain the policy from four complementary aspects, 
including pose consistency, velocity matching, dynamic motion trend tracking, 
and key support joint stabilization. This design allows the robot to preserve the periodic structure imposed by the reference gait generator while maintaining adaptability to complex environments.

\begin{table}[t]
\caption{GAIT-AWARE REWARD COMPONENTS}
\label{tab:gait_rewards}
\centering
\renewcommand{\arraystretch}{1.2}

\begin{tabular*}{\columnwidth}{@{\extracolsep{\fill}}lcc}
\hline
\textbf{Reward} & \textbf{Equation ($e_i$)} & \textbf{Weight ($w_i$)} \\
\hline
$r_{\text{pos}}$ 
& $\| \boldsymbol{\theta} - \boldsymbol{\theta}_d(\phi) \|^2$ 
& 0.10 \\
$r_{\text{vel}}$ 
& $\| \mathbf{v}_b - \mathbf{v} \|^2$ 
& 0.05 \\
$r_{\Delta}$ 
& $\| \Delta \boldsymbol{\theta} - \Delta \boldsymbol{\theta}_d(\phi) \|_1$ 
& 0.05 \\
$r_{\text{ankle}}$ 
& $\sum_{j \in \{L,R\}} 
\left( \theta_j - \theta_{d,j}(\phi) \right)^2$ 
& 0.05 \\
\hline
\end{tabular*}

\end{table}

\subsection{High-Throughput Training Infrastructure}
\label{sec:training_optimization}
%To address the computational overhead and GPU memory (VRAM) bottlenecks caused by high-dimensional depth perception in massively parallel environments, we developed a systematic engineering optimization framework on the NVIDIA Isaac Sim and Isaac Lab platform. To the best of our knowledge, this is the first work within the Isaac Lab framework to achieve complex terrain training for humanoid robots using a single depth camera as the sole visual input. Based on our self-developed ZERITH Z1 humanoid robot model, we constructed the training environment and improved overall system throughput from three perspectives: pipeline parallelism, memory management, and rendering strategy.

To address the computational overhead and GPU memory (VRAM) bottlenecks caused by high-dimensional depth perception in massively parallel environments, we developed a systematic engineering optimization framework on the NVIDIA Isaac Sim and Isaac Lab platform. Based on our self-developed ZERITH Z1 humanoid robot model, we constructed the training environment and improved overall system throughput from two perspectives: memory management and rendering strategy.

% \subsubsection{Zero-copy Perceptual Pipeline via NVIDIA Warp}
% Traditional visual data pipelines are typically constrained by frequent Python wrapper invocations and redundant memory copies, which lead to significant I/O latency when scaling to thousands of parallel environments. To overcome this limitation, we redesigned a high-performance image reorganization pipeline using NVIDIA Warp. By implementing custom GPU kernels, the system performs tiled image reconstruction and splitting directly in VRAM in a fully parallel manner. Furthermore, through memory aliasing, we aligned VRAM addresses between the simulator and the deep learning framework, eliminating unnecessary data transfers. This near-zero-overhead observation handoff substantially reduces end-to-end perception latency.

\subsubsection{Heterogeneous Observation Buffer Management}
VRAM capacity is a primary factor limiting the degree of parallelism (i.e., $N_{\mathrm{env}}$) in reinforcement learning. To address the high-dimensional tensor storage pressure introduced by depth images, we propose a heterogeneous memory management scheme designed to decouple physics computation from data caching. Under this mechanism, VRAM serves only as a transient buffer for rendering outputs, while the generated observation tensors are asynchronously transferred to host memory (RAM) for storage and indexing. 

Experimental results demonstrate that this strategy significantly frees GPU computational resources. As shown in Table~\ref{tab:memory_parallel}, on an RTX~4090 (24\,GB) platform, the maximum number of parallel environments for vision-based tasks increases from the baseline of 512 to 1024. In a 48\,GB VRAM configuration, the parallel scale further extends to 1536 environments, substantially improving sample efficiency during training.

\begin{table}[t]
\caption{PARALLEL SCALE UNDER DIFFERENT MEMORY STRATEGIES}
\label{tab:memory_parallel}
\centering
\renewcommand{\arraystretch}{1.2}

\begin{tabular*}{\columnwidth}{@{\extracolsep{\fill}}lccc}
\hline
\textbf{GPU} & \textbf{Max $N_{\mathrm{env}}$} & \textbf{Storage} & \textbf{VRAM} \\
\hline
4090 (24G) & 512  & GPU & $\sim$24G \\
4090 (24G) & 1024 & CPU & $\sim$24G \\
4090 (48G) & 1536 & CPU & $\sim$48G \\
\hline
\end{tabular*}
\end{table}

\subsubsection{Render-time Pre-processing Optimization}
To optimize the vision processing pipeline, we first define the core perceptual requirement: the system must reliably distinguish terrain features with a minimum height of $5\,\text{cm}$. We derive the spatial resolution based on the geometric camera model:
\begin{equation}
\text{r}_v = \frac{z_0 \sin(-\delta)}{\sin(\alpha) \sin(\alpha + \delta)} = \frac{z_0 \sin(\frac{\beta}{h})}{\sin(\alpha) \sin(\alpha - \frac{\beta}{h})}
\end{equation}
where $z_0$ represents the mounting height, $\alpha$ is the pitch angle, $\beta$ denotes the vertical Field of View (FOV), $h$ is the image height in pixels, and auxiliary variable $\delta = -\beta / h$.

Under the configuration of $z_0 = 0.8\,\text{m}$, $\beta = 58^\circ$, and $\alpha = 45^\circ$, an effective vertical resolution of $h=36\,\text{px}$ yields a spatial resolution of $r_v$ = $0.0463\,\text{m/pix}$ at a typical measurement distance of $1.13\,\text{m}$. Since $0.0463\,\text{m} < 0.05\,\text{m}$, this configuration satisfies the critical constraint for sensing $5\,\text{cm}$ terrain variations while minimizing the input dimensionality.

\begin{figure}[t]
   \centering
   \includegraphics[width=1.0\columnwidth]{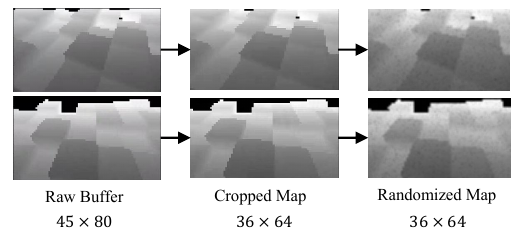}
   \caption{Depth data flow: Images in the top and bottom rows represent samples from different parallel environments.}
   \label{fig:depth_image_fig}
\end{figure}

Based on this analysis, we directly render a low-resolution depth buffer of $45 \times 80$ instead of high-resolution frames. A $36 \times 64$ region is extracted via center cropping, followed by stochastic depth perturbations to improve robustness. This render-time optimization reduces perception overhead and GPU memory bandwidth usage, enabling the policy to process visual observations at a significantly higher frequency, which is crucial for maintaining stable locomotion on highly irregular terrains.. The overall pipeline is illustrated in Fig.~\ref{fig:depth_image_fig}.

\subsection{Training Details}
% \subsubsection{Simulation Platform}
% Building upon the optimization framework described in Section~\ref{sec:training_optimization}, we achieved high-throughput policy training on a single NVIDIA RTX~4090 (24\,GB) GPU. The system reliably supports 1024 parallel agents performing depth perception and physics simulation simultaneously. The policy converges after approximately 40{,}000 training iterations, with a total wall-clock time of only 20 hours. The trained policy is exported via ONNX and can be directly deployed to the onboard computing unit of the ZERITH Z1 humanoid robot.

\subsubsection{Training Curriculum}
\begin{figure}[t]
    \centering
    \scriptsize 
    
    % 第一行地形：A 和 B
    \begin{minipage}{0.49\columnwidth}
        \centering
        \begin{overpic}[width=\textwidth]{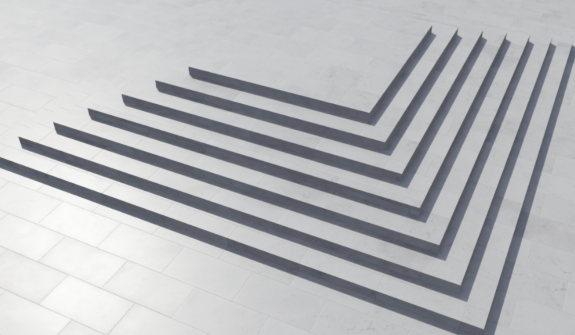}
            % (3, 88) 坐标：3是距离左边3%，53是距离底边53%（即左上角）
            \put(2,53){\textbf{A}} 
        \end{overpic}
    \end{minipage}
    \hfill
    \begin{minipage}{0.49\columnwidth}
        \centering
        \begin{overpic}[width=\textwidth]{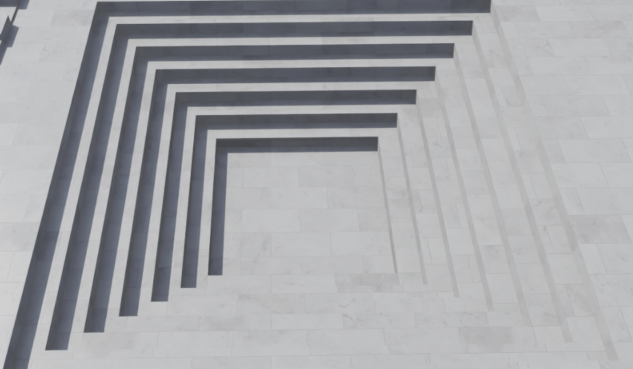}
            \put(3,53){\textbf{B}}
        \end{overpic}
    \end{minipage}

    \vspace{3pt} 

    % 第二行地形：C 和 D
    \begin{minipage}{0.49\columnwidth}
        \centering
        \begin{overpic}[width=\textwidth]{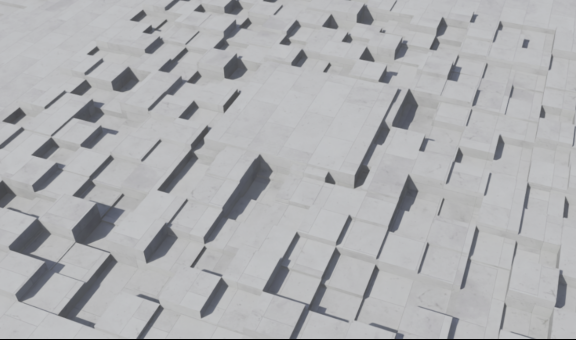}
            \put(3,53){\textbf{C}}
        \end{overpic}
    \end{minipage}
    \hfill
    \begin{minipage}{0.49\columnwidth}
        \centering
        \begin{overpic}[width=\textwidth]{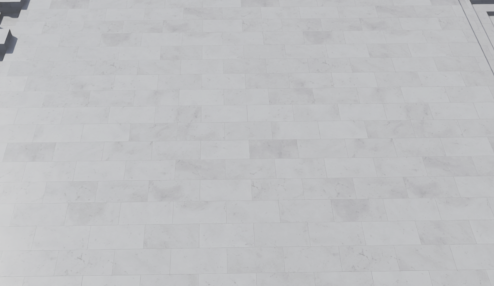}
            \put(3,53){\textbf{D}}
        \end{overpic}
    \end{minipage}

    \vspace{6pt} 
    % 严格遵循你要求的表格格式
    \centering
    \scriptsize 
    \renewcommand{\arraystretch}{1.2}
    \begin{tabular*}{\columnwidth}{@{\extracolsep{\fill}}lcccc}
        \hline
        \textbf{E} & \textbf{A} & \textbf{B} & \textbf{C} & \textbf{D} \\
        \hline
        Terrain Type & Pyramid Stairs & Inverted Stairs & Boxes & Plane \\
        Range (/m) & $[0.05, 0.23]$ & $[0.05, 0.23]$ & $[0.05, 0.20]$ & -- \\
        Parameter & step height & step height & obstacle height & flat surface \\
        Weight & 0.2 & 0.2 & 0.2 & 0.1 \\
        \hline
    \end{tabular*}

    \caption{Overview of the terrain curriculum for training. (A)--(D) represent distinct terrain types posing various physical challenges. Table (E) details the parameter ranges and the distribution (weight) for each terrain type.}
    \label{fig:terrain_config}
\end{figure}

To prevent policy instability caused by overly challenging terrains during the early stage of training, we adopt an adaptive terrain curriculum strategy based on the traveled distance of the robot, following the approach in \cite{DBLP:conf/corl/RudinHR021}. In simulation, curriculum training is conducted simultaneously over four types of terrain: Pyramid Stairs, Inverted Stairs, Boxes, and Plane. All four terrain types are native terrain generators provided by Isaac Lab. Detailed terrain configurations and curriculum settings are illustrated in Fig.~\ref{fig:terrain_config}.

\subsubsection{Domain Randomization}
To enhance robustness and sim-to-real transfer, we apply domain randomization across physical dynamics, initial conditions, and visual inputs. The corresponding perturbation ranges are listed in Table \ref{tab:dr_param}.

% \subsubsection{Random Seed Settings}
% To ensure the reproducibility of the experimental results, we set fixed random seeds for all randomized components, including policy initialization, terrain generation, and domain randomization. The experiment was conducted under consistent random conditions.

\begin{table}[t]
\caption{DOMAIN RANDOMIZATION RANGES}
\label{tab:dr_param}
\centering
\renewcommand{\arraystretch}{1.2}
\begin{tabular*}{\columnwidth}{@{\extracolsep{\fill}}lcc}
\hline
\textbf{Parameter} & \textbf{Randomization range} & \textbf{Unit} \\
\hline
Base payload & $[-5.0, 5.0]$ & kg \\
Link mass factor & $[0.8, 1.2]$ & -- \\
Center of mass shift & $[-0.15, 0.15]$ & m \\
Friction coefficient & $[0.2, 1.5]$ & -- \\
$K_p$ factor & $[0.9, 1.1]$ & -- \\
$K_d$ factor & $[0.9, 1.1]$ & -- \\
Joint armature & $[2\times10^{-3}, 2\times10^{-2}]$ & kg$\cdot$m$^2$ \\
Initial base position ($x, y$) & $[-0.5, 0.5]$ & m \\
Initial base orientation (yaw) & $[-\pi, \pi]$ & rad \\
Initial base linear velocity & $[-0.5, 0.5]$ & m/s \\
Initial base angular velocity & $[-0.5, 0.5]$ & rad/s \\
Initial joint position scale & $[0.5, 1.5]$ & -- \\
Depth image bias & $[-0.04, 0.04]$ & m \\
Depth image noise ($\sigma$) & $0.02$ & m \\
Depth hole probability & $0.03$ & -- \\
\hline
\end{tabular*}
\end{table}
%%%%%%%%%%%%%%%%%%%%%%%%%%%%%%%%%%%%%%%%%%%%%%%%%%%%%
\section{EXPERIMENT}

\begin{table*}[t]
\caption{ABLATION STUDY RESULTS OF THE PRIOR FRAMEWORK ON TERRAIN ADAPTABILITY}
\label{tab:ablation_results}
\centering
\renewcommand{\arraystretch}{1.2}
\begin{tabular*}{\textwidth}{@{\extracolsep{\fill}}lcccccc}
\hline
\textbf{Method} & \textbf{Mean Level} & \textbf{Pyramid Stairs} & \textbf{Inverted Stairs} & \textbf{Boxes} & \textbf{Plane} & \textbf{Mean Reward} \\
\hline
PRIOR (ours) & 5.7533 & 1.0 & 1.0000 & 1.0000 & 1.0 & \textbf{26.3462} \\
PRIOR w/o reference gait & \textbf{5.7735} & 1.0 & 1.0000 & 1.0000 & 1.0 & 23.7233 \\
PRIOR w/o $\hat{\mathbf{m}}_t$ & 5.7672 & 1.0 & 0.7734 & 1.0000 & 1.0 & 13.1775 \\
PRIOR w/o $\mathbf{d}_t^{\mathrm{H2}}$ & 5.4627 & 1.0 & 0.3750 & 0.9687 & 1.0 & 10.1463 \\
PRIOR with H1 = 6 & 5.7417 & 1.0 & 1.0000 & 1.0000 & 1.0 & 19.3234 \\
PRIOR w/o landing state reward & 5.7403 & 1.0 & 1.0000 & 1.0000 & 1.0 & 22.6262 \\
\hline
\end{tabular*}
\end{table*}

\subsection{Experimental Setting}
Based on the optimization architecture described in Section~\ref{sec:training_optimization}, we conduct high-throughput policy training on a single NVIDIA RTX 4090 (24GB) GPU. The system supports 1024 parallel environments simultaneously performing depth perception and physics simulation, significantly improving data collection efficiency while maintaining stable simulation dynamics. The policy converges after approximately 12,000 training iterations.

The depth images are updated at 30 Hz, while the control policy is executed at 50 Hz, ensuring a balance between perception refresh rate and control stability.

The trained policy is exported via ONNX and deployed directly on the onboard computing unit of the ZERITH Z1 humanoid robot. The ZERITH Z1 model has 23 degrees of freedom (DoF), including 6 DoF per leg, 3 DoF at the waist, and 4 DoF per arm.

\subsection{Simulation Ablation Studies}
\subsubsection{Experimental Configurations}
To systematically analyze the contribution of each component in the PRIOR framework, we design a staged ablation study to separately evaluate the perception-driven locomotion architecture and the reference gait prior.

First, four ablated variants are compared against the configuration without the reference gait prior (``PRIOR w/o reference gait'') to investigate the impact of architectural components within the perception-motion framework. Subsequently, the ``PRIOR w/o reference gait'' model is compared with the full PRIOR framework to quantify the contribution of the reference gait prior to locomotion stability and behavioral quality.

The detailed configurations are as follows:

\begin{itemize}
    \item \textbf{PRIOR (Ours)}: The complete PRIOR framework.
    \item \textbf{PRIOR w/o reference gait}: The PRIOR framework without the humanoid reference gait constraint.
    \item \textbf{PRIOR w/o $\hat{\mathbf{m}}_t$}: The PRIOR framework without explicit terrain estimation and elevation map supervision.
    \item \textbf{PRIOR w/o $\mathbf{d}_t^{\mathrm{H2}}$}: The PRIOR framework where temporally stacked depth observations are removed from the estimator input.
    \item \textbf{PRIOR with shorter H1}: The PRIOR framework with a reduced proprioceptive history length (H1 = 6).
    \item \textbf{PRIOR w/o landing state reward}: The PRIOR framework without the designed landing state reward terms.
\end{itemize}

\subsubsection{Evaluation Procedure}

All six policies are trained in the Isaac Lab simulator under the same terrain curriculum setting. We evaluate the policies using the following quantitative metrics:

\begin{itemize}
    \item \textbf{Curriculum capability}: The mean terrain level achieved (mean level) and the maximum successfully traversed level across different terrain types (Pyramid Stairs, Inverted Stairs, Boxes, and Plane).
    \item \textbf{Training performance}: Total reward, convergence speed, and training stability.
\end{itemize}

%%%%%%%%%%%%%%%%%%%%%%%%%%%%%%%%%%%%%%%%%%%%%%%
\begin{figure}[t]
   \centering
   \includegraphics[width=1.0\columnwidth]{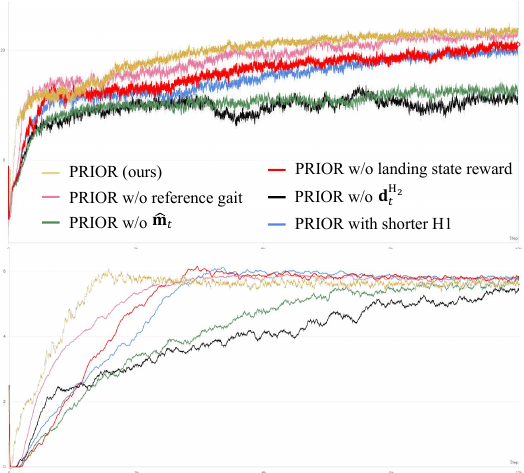}
   \caption{The upper panel illustrates the training curves for mean reward, while the lower panel displays the mean terrain level achieved over training iterations.}
   \label{fig:curve_fig}
\end{figure}
%%%%%%%%%%%%%%%%%%%%%%%%%%%%%%%%%%%%%%%%%%%%%%%

\subsubsection{Results and Discussion}
As shown in Table~\ref{tab:ablation_results} and Fig.~\ref{fig:curve_fig}, the ablation results demonstrate the effectiveness of each component.

\textbf{Reward–stability correlation}:  
Although the ``PRIOR w/o reference gait'' variant achieves a slightly higher mean curriculum level (5.7735), its average reward (23.7233) is approximately 10\% lower than that of the full framework. This suggests that traversal capability alone does not guarantee motion quality or efficiency.

\textbf{Behavioral rationality}:  
The reference gait prior contributes not only to performance but also to motion quality. Empirically, policies without gait prior tend to exploit unstable or high-frequency oscillatory motions to maximize traversal success. With the humanoid gait constraint, the robot maintains near-perfect terrain success rates (all terrain metrics reaching 1.0) while achieving significantly smoother and more energy-efficient locomotion, as reflected by higher reward values. This property is crucial for real-world deployment.

\textbf{Explicit terrain estimation and supervision (PRIOR w/o $\hat{\mathbf{m}}_t$)}: 
Removing explicit terrain estimation reduces the average reward to 13.1775. Although it performs better than the variant without temporal depth information, its performance on complex terrains such as Inverted Stairs (0.7734) remains substantially lower than the full model. This indicates that explicit terrain representation and elevation supervision significantly enhance the policy’s understanding of geometric structures.

\textbf{Temporal depth information (PRIOR w/o $\mathbf{d}_t^{\mathrm{H2}}$)}:  
This variant exhibits the worst performance, with an average reward of 10.1463 and a significant drop in success rate on Inverted Stairs (0.375). Without integrating temporal depth information, the robot loses the ability to anticipate terrain variations, leading to unstable foothold planning during dynamic locomotion.

\textbf{Proprioceptive history length (H1)}:  
Reducing the proprioceptive history length results in an average reward of 19.3234. A longer history window (H1 = 10) enables the system to implicitly estimate physical properties such as ground friction and center-of-mass deviation, thereby improving robustness under unknown disturbances.

\begin{figure}[t]
   \centering
   \includegraphics[width=1.0\columnwidth]{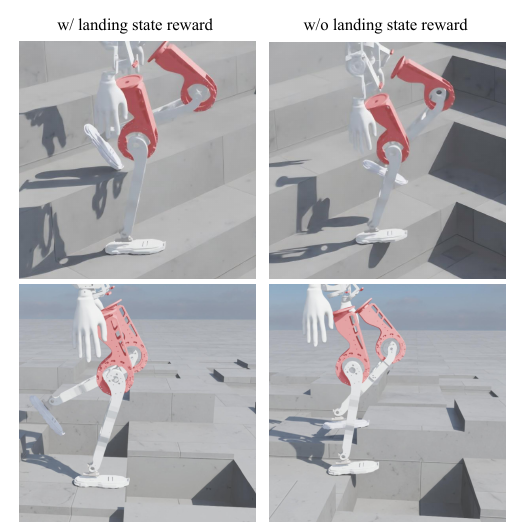}
   \caption{Comparison of foot-placement behavior with and without the landing state reward.}
   \label{fig:landingstate_fig}
\end{figure}

\textbf{Landing state reward}:  
As illustrated in Fig.~\ref{fig:landingstate_fig}, the removal of the landing state reward results in less stable foot placements, leading to a decrease in mean reward to 22.6262. This qualitative comparison reinforces that our fine-grained reward design effectively optimizes the behavior during the landing transient.
%%%%%%%%%%%%%%%%%%%%%%%%%%%%%%%%%%%%%%%%%%%%%%%
%%%%%%%%%%%%%%%%%%%%%%%%%%%%%%%%%%%%%%%%%%%%%%%
%%%%%%%%%%%%%%%%%%%%%%%%%%%%%%%%%%%%%%%%%%%%%%%%%%%
\section{CONCLUSIONS}
This paper presents \textbf{PRIOR}, an efficient and reproducible single-stage reinforcement learning framework developed in the Isaac Lab environment for perception-aware humanoid locomotion. The proposed method addresses the challenges of integrating perception and control for humanoid robots operating over complex terrains. \textbf{PRIOR} combines a GRU-based explicit terrain reconstruction state estimator with a parameterized gait generator within a unified learning pipeline. This design enables the ZERITH Z1 humanoid model to traverse diverse challenging terrains with high precision and robustness. Experimental results demonstrate that PRIOR achieves high average rewards and maintains a 100\% traversal success rate across various complex terrains. Furthermore, the learned policy consistently handles terrains of relatively high difficulty, highlighting the effectiveness and generalization capability of the proposed framework.

Despite these advantages, our current work still has limitations in real-world deployment experiments. Future research will focus on developing more generalizable dynamics adaptation algorithms to enable seamless sim-to-real transfer from Isaac Lab to the physical ZERITH Z1 platform.

\addtolength{\textheight}{-12cm}   % This command serves to balance the column lengths

%%%%%%%%%%%%%%%%%%%%%%%%%%%%%%%%%%%%%%%%%%%%%%%%%%%%%%%%%%%%%%%%%%%%%%%%%%%%%%%%

%%%%%%%%%%%%%%%%%%%%%%%%%%%%%%%%%%%%%%%%%%%%%%%%%%%%%%%%%%%%%%%%%%%%%%%%%%%%%%%%



% \section*{ACKNOWLEDGMENT}

% The preferred spelling of the word ÒacknowledgmentÓ in America is without an ÒeÓ after the ÒgÓ. Avoid the stilted expression, ÒOne of us (R. B. G.) thanks . . .Ó  Instead, try ÒR. B. G. thanksÓ. Put sponsor acknowledgments in the unnumbered footnote on the first page.

%%%%%%%%%%%%%%%%%%%%%%%%%%%%%%%%%%%%%%%%%%%%%%%%%%%%%%%%%%%%%%%%%%%%%%%%%%%%%%%%

\bibliographystyle{IEEEtran}

\bibliography{ref}

\end{document}